\documentclass[11pt, letterpaper]{googledeepmind}

\usepackage{multirow}
\usepackage[numbers, sort&compress]{natbib}
\usepackage{lipsum}
\usepackage{amsmath}
\usepackage{pstricks, pst-node}
\usepackage{verbatim}
\usepackage{multicol}
\usepackage{longtable}
\usepackage{array}
\usepackage{listings, listings-rust}
\usepackage{enumitem}
\usepackage{xspace}
\usepackage{bm}
\usepackage{bbm}
\usepackage{mathtools}
\usepackage{soul}
\usepackage{graphicx}
\usepackage[most, breakable, skins]{tcolorbox}
\tcbuselibrary{skins}
\usepackage{subcaption}
\usepackage{amssymb}
\usepackage{colortbl}
\usepackage{csquotes}
\usepackage{setspace}
\usepackage[inkscapeformat=png]{svg}
\usepackage{tabularx,ragged2e}
\usepackage{makecell}
\usepackage{placeins}
\usepackage[symbol]{footmisc}
\usepackage[dvipsnames]{xcolor}
\usepackage{hyperref}
\usepackage{pifont}
\usepackage{cleveref}
\usepackage{fontawesome}
\usepackage{float}
\usepackage{microtype}  

\usepackage{pdfpages}

\newcolumntype{L}[1]{>{\raggedright\let\newline\\\arraybackslash\hspace{0pt}}m{#1}}
\newcolumntype{C}[1]{>{\centering}m{#1}}

\newcolumntype{R}[1]{>{\raggedleft\let\newline\\\arraybackslash\hspace{0pt}}m{#1}}

\definecolor{ao}{rgb}{0.0, 0.0, 1.0}

\newcommand{\todo}[1]{}
\newcommand{\todor}[1]{}

\newcommand{\optimam}[1]{}

\newcommand\vcent[1]{\vcenter{\hbox{#1}}}
\newcommand\loudspeaker[1][3]{\ensuremath{\vcent{\rule{.6ex}{.6ex}}\kern-.5ex%
  \vcent{\scalebox{.6}[1]{\rotatebox[origin=center]{90}{$\blacktriangle$}}}%
  \ifnum#1>0\relax\kern.05ex\vcent{\scalebox{.4}{\ttfamily)}}%
  \ifnum#1>1\relax\kern-.4ex\vcent{\scalebox{.56}{\ttfamily)}}%
  \ifnum#1>2\relax\kern-.55ex\vcent{\scalebox{.7}{\ttfamily)}}%
  \fi\fi\fi}%
}

\title{Towards a Medical AI Scientist}

\author{
  Hongtao Wu\textsuperscript{1,*} \quad
  Boyun Zheng\textsuperscript{1,*} \quad
  Dingjie Song\textsuperscript{2,*} \quad
  Yu Jiang\textsuperscript{1} \quad
  Jianfeng Gao\textsuperscript{4} \quad
  Lei Xing\textsuperscript{3} \quad
  Lichao Sun\textsuperscript{2,\dag} \quad
  Yixuan Yuan\textsuperscript{1,\dag} \\
  \small \textsuperscript{1}The Chinese University of Hong Kong \quad
  \textsuperscript{2}Lehigh University \quad
  \textsuperscript{3}Stanford University \quad
  \textsuperscript{4}Microsoft Research \\
  \small \texttt{\{lis221@lehigh.edu, yxyuan@ee.cuhk.edu.hk\}} \\
  \faGithub~\textbf{Homepage:} \textcolor{blue}{\url{https://cuhk-aim-group.github.io/Med-AI-Scientist-Homepage/}}
}

\begin{abstract}
Autonomous systems that generate scientific hypotheses, conduct experiments, and draft manuscripts have recently emerged as a promising paradigm for accelerating discovery.
However, existing “AI Scientists” remain largely domain-agnostic, limiting their applicability to clinical medicine, where research is required to be grounded in medical evidence with specialized data modalities.
In this work, we introduce Medical AI Scientist, the first autonomous research framework tailored to clinical autonomous research.
It generates clinically grounded ideas by transforming surveyed literature into actionable evidence through a clinician-engineer co-reasoning mechanism, which improves the traceability of generated research ideas.
The Medical AI scientist further introduces evidence-grounded manuscript drafting guided by a structured medical writing paradigm and ethical policies.
The framework operates under 3 research modes, namely paper-based reproduction, literature-inspired innovation, and task-driven exploration, corresponding to distinct levels of medical scientific autonomy.
Comprehensive evaluations by both large language models and human experts demonstrate that the ideas generated by the Medical AI Scientist are of substantially higher quality than those produced by commercial LLMs across 171 cases, covering 19 clinical tasks, and 6 data modalities.
Meanwhile, our system achieves strong alignment between the proposed method and its implementation, while also demonstrating significantly higher success rates in executable experiments.
Double-blind evaluations by human experts and the Stanford Agentic Reviewer suggest that the generated manuscripts approach MICCAI-level quality, while consistently surpassing those from ISBI and BIBM.
The proposed Medical AI Scientist highlights the potential of leveraging AI for autonomous scientific discovery in healthcare.  
\end{abstract}

\renewcommand{\copyrightext}{}

\begin{document}

\maketitle

\section{Introduction}
\label{sec:intro}
Recent years have witnessed rapid advances in artificial intelligence for healthcare, with increasingly capable models achieving state-of-the-art performance across disease diagnosis~\cite{esteva2017dermatologist,kermany2018identifying,rajpurkar2017chexnet,zhang2023large}, medical image analysis~\cite{isensee2018nnunet,hatamizadeh2022unetr,ma2024segment} and clinical outcome prediction~\cite{mobadersany2018predicting,wang2024pathology,chen2024metabolomic}. In parallel, large language models~\cite{GPT-5, gpt-oss, team2023gemini, Grok4, qwen3, deepseek} have made substantial progress in language understanding, reasoning and code generation, enabling the emergence of tool-augmented and multi-agent systems~\cite{openai-deepresearch, gemini-deepresearch, xai-deepresearch, autogen, camel, metagpt, xie2023OpenAgents, manus2025, openmanus2025} that extend beyond narrow task execution. Together, these developments have catalyzed the rise of autonomous research frameworks, often referred to as AI Scientists~\cite{ai-scientist, ai-scientistv2, AI-researcher,deepscientist}, which seek to automate the scientific workflow from hypothesis generation and experimental design to result interpretation and manuscript preparation, promising to accelerate scientific innovation~\cite{reddy2025towards}. These AI Scientist systems have shown promise in accelerating research in domains such as mathematics, chemistry and general machine learning, where problem formulations, data representations and evaluation protocols are relatively standardized. 

Medical AI represents one of the most consequential domains for such systems, given its direct implications for patient outcomes, diagnostic reliability and healthcare efficiency. As medical datasets, analytical methodologies and scientific literature continue to grow at an unprecedented pace, the throughput of human-driven research has become an increasingly critical bottleneck~\cite{gil2014amplify, wang2023scientific, baek2024researchagent,gottweis2025towards}. This widening gap highlights the urgent need for autonomous scientific systems that are explicitly designed to operate within the epistemic, operational, and ethical constraints inherent to clinical medicine.

However, extending these autonomous research paradigms to medical field remains challenging.
First, existing AI Scientists focus on model modifications or generic optimization strategies, ignoring medical related priors, such as basic diagnostic workflows and disease-specific pathological patterns. 
Moreover, their retrieval and reasoning processes frequently lack sufficient constraints to reliably identify authoritative medical reasoning evidence, which will lead to models with superficial performance metrics but fail to capture clinically relevant patterns.
Secondly, the heterogeneous and high dimensional nature of medical data, including three dimensional and anisotropic structures, together with specialized evaluation standards, poses challenges to the reliable and fair experimentation execution. 
Thirdly, the provenance of medical data and the clarity of ethical statements are central to the credibility, reproducibility, and clinical translation of research findings, yet current autonomous research systems largely overlook these requirements and fail to produce manuscripts that adhere to clinical writing frameworks and ethical standards.

\begin{figure}
    \centering
    \includegraphics[width=1\linewidth]{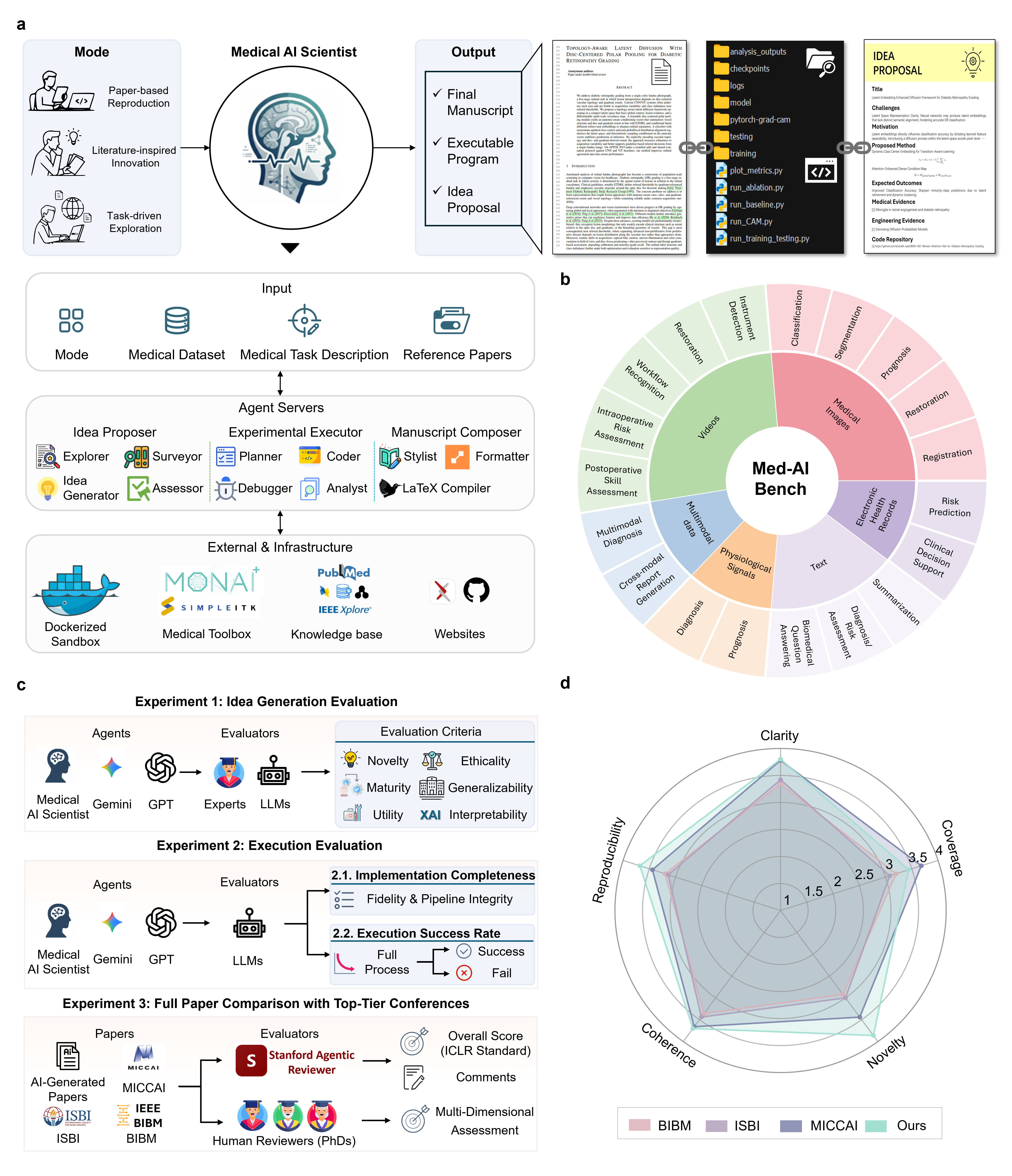}
    \caption{
    a, System workflow: fully-automated multi-agents system for end-to-end scientific discovery in clinical medicine.
    b, Med-AI Bench: visualization depicting 19 distinct medical research tasks within performance benchmarking.
    c, Experimental setup: comparative evaluation across Idea generation, execution and full paper compilation in the research lifecycle.
    d, Performance benchmarking: comparable manuscript quality to representative works from leading venues.
    }
    \label{fig:framework}
\end{figure}

Here we present Medical AI Scientist, an agentic framework for end-to-end medical AI discovery and development, as shown in Fig.~\ref{fig:framework}a. 
The system comprises three key components: Idea Proposer,  Experimental Executor, and Manuscript Composer, which together support the fully autonomous research lifecycle.
The Idea Proposer leverages structured literature retrieval and analysis to identify clinical prior and then adapts the most suitable emerging technical models to medical tasks. 
A clinician–engineer co-reasoning mechanism is incorporated into the idea generation process to explicitly ground each hypothesis in verifiable evidence and mitigate hallucinations.
The automated experimental executor orchestrates a reliable validation pipeline by unifying general-purpose execution toolchains with domain-specific medical toolboxes tailored to heterogeneous and complex clinical data formats, enabling iterative and self-correcting deep model development.
A hierarchical Manuscript Composer transforms research outputs into coherent and evidence-grounded drafts through a structured medical writing paradigm with enhanced narrative logic and readability. It also embeds ethical review mechanisms that explicitly document data usage in compliance with medical publication policies.

To address the absence of standardized evaluation protocols for automated medical research systems, we introduce Med-AI Bench (Fig.~\ref{fig:framework}b). 
This benchmark comprises 171 high-quality evaluation cases, organized around 19 distinct research tasks spanning 6 common medical data modalities. 
For each task, we selected 3 representative papers of varying difficulty (easy, medium, hard) and constructed 3 evaluation cases with different input modes. This design provides a systematic and unified framework for both qualitative and quantitative assessment of automated medical research systems across the full research pipeline.

As presented in Fig.~\ref{fig:framework}c,
we first evaluate research idea generation using both large language models and human experts (Fig.~\ref{fig:fig3}), showing that the Medical AI Scientist consistently surpasses commercial language models across six dimensions, including novelty, maturity, ethicality, generalizability, utility, and interpretability. We then assess experimental execution, where the system exhibits strong alignment between proposed methods and their implementations, together with substantially higher success rates in producing executable experiments (Fig.~\ref{fig:figure_2}). Finally, under double blind evaluation (Fig.~\ref{fig:framework}d,~\ref{fig:fig4}b \& c), 10 independent domain experts assess generated manuscripts alongside high quality human authored studies from leading venues such as MICCAI, ISBI, and BIBM, while all submissions were further reviewed using the Stanford Agentic Reviewer under ICLR-aligned criteria (Fig. ~\ref{fig:fig4}a). The generated manuscripts achieve a mean score of 4.60 ± 0.56 and remain competitive across key dimensions including novelty, reproducibility, coherence, and clarity, with only a modest gap in coverage. Qualitative feedback further indicates strong practical relevance and clear presentation with limited critical weaknesses. 
Moreover, one manuscript generated by our system has been accepted by the International Conference on AI Scientists (ICAIS~2025~\citep{icais2025}) after peer review.
Together, these results suggest that automated systems can speed up complex methodological designs, highlighting their potential to significantly enhance the efficiency of medical AI research.

\section{Results}
\label{sec:Results}

\subsection{Building universal medical research by systematic LLM Agent} 

The Medical AI Scientist provides different levels of autonomous academic research modes: Paper-based Reproduction, Literature-inspired Innovation, and Task-driven Exploration.
These modes are designed to accommodate users ranging from early stage PhD-level researchers entering a medical AI task to domain experts seeking efficient and highly automated solutions for open ended problems.
The Reproduction mode follows explicitly defined research instructions derived from target papers and focuses on the faithful implementation of established methods. An ethical gatekeeping mechanism is incorporated to prevent harmful implementations.
Instead of relying on explicit method specifications, 
the Innovation mode identifies research gaps and generates hypotheses based on fixed references and datasets. Evaluation emphasizes originality and methodological completeness, supported by a clinician–engineer co-reasoning mechanism and multi-dimensional assessment.
The Exploration mode further targets problem driven discovery in real-world settings. Starting from a single user defined question, the system conducts literature mining, selects and integrates paradigms, generates solutions, and performs experimental verification.

To enable a rigorous and domain-spanning assessment of the Medical AI Scientist, we constructed Med-AI Bench, a benchmark grounded in peer-reviewed medical AI literature and expert-annotated references. Med-AI Bench is deliberately organized to reflect the breadth of contemporary medical AI research, covering six data modalities and nineteen representative tasks that span the full spectrum from low-level perception to high-level clinical reasoning (Fig. \ref{fig:framework}b).
Specifically, medical images-related tasks cover core problems in visual understanding and analysis, including classification~\cite{manzari2023medvit,huo2024hifuse,yang2025diffmic}, segmentation~\cite{ronneberger2015u,cao2022swin,chen2024transunet}, prognosis~\cite{hermoza2021post,chato2017machine,lin2025glioblastoma}, registration~\cite{balakrishnan2019voxelmorph,chen2022transmorph,kim2022diffusemorph}, and restoration~\cite{wang2021dicdnet,wang2023oscnet,wang2023mepnet}. Video-centric tasks encompass instrument detection~\cite{wang2024video,wu2023onlinerefer,botach2022end}, 
restoration~\cite{liu2025medvsr, wang2019edvr,chan2021basicvsr},
workflow recognition~\cite{yang2024surgformer, jin2021temporal, jin2017sv},  intraoperative risk assessment~\cite{kawamura2023development,mascagni2022artificial,nowak2025swincvs}, and postoperative skill assessment~\cite{zia2018automated,funke2019video,liu2021towards}. 
Structured electronic health record data support tasks in risk prediction~\cite{im2025labtop,poulain2024graph,fallahpour2024ehrmamba} and clinical decision support~\cite{sun2023cehmr, shang2019gamenet,yang2021safedrug}, while physiological signal data are used for diagnosis~\cite{el2024ecgtransform,wang2024medformer,yang2023multi} and prognosis~\cite{lima2021deep,raghunath2020prediction,sangha2022automated}. Text-based clinical reasoning is evaluated through report summarization~\cite{van2024adapted,yadav2021reinforcement,lu2024medical}, diagnosis and risk assessment~\cite{huang2019clinicalbert,golmaei2021deepnote,ma2024hr}, and biomedical question answering~\cite{wiese2017neural,yang2016learning,kim2025prompting}. Finally, multimodal tasks assess the system’s ability to integrate heterogeneous data sources for multimodal diagnosis~\cite{zhang2023tformer,zhang2025novel,cockayne2025dermformer} and cross-modal report generation~\cite{yang2022knowledge,hou2023recap,chen2020generating}.

For each task, we retrieve three papers from Google Scholar, which serve as a structured ground truth for benchmarking different levels of scientific reasoning and execution. 
Each paper was evaluated across five dimensions, including code availability, venue quality, citations, year, complexity, and subjective human rating, and then ranked and assigned to one of three difficulty tiers per task.
Using this benchmark, we evaluate the Medical AI Scientist across the complete research lifecycle, including idea generation, experimental execution, and manuscript compilation. 
Collectively, Med-AI Bench functions as a standardized and reproducible framework for assessing autonomous medical AI researchers under realistic, multi-modal, and clinically relevant research conditions.

\subsection{Comprehensive evaluation of idea generation}

The Idea Generation module is designed to address two central challenges in AI assisted research ideation. The first concerns the generation of novel hypotheses from unstructured resources without a specific direction, as in the Innovation mode. The second concerns the need to ensure that these hypotheses remain clinically relevant and technically feasible, which is emphasized in the Exploration mode.
We quantitatively evaluated the quality of model-generated research ideas against two commercial LLMs (e.g., GPT-5, Gemini-2.5-Pro), using both LLM-as-judge metrics and blinded human assessments, with evaluations conducted across six criteria commonly adopted in medical AI research, including novelty, maturity, ethicality, generalizability, utility, and interpretability.

\begin{figure}
    \centering
    \includegraphics[width=0.95\linewidth]{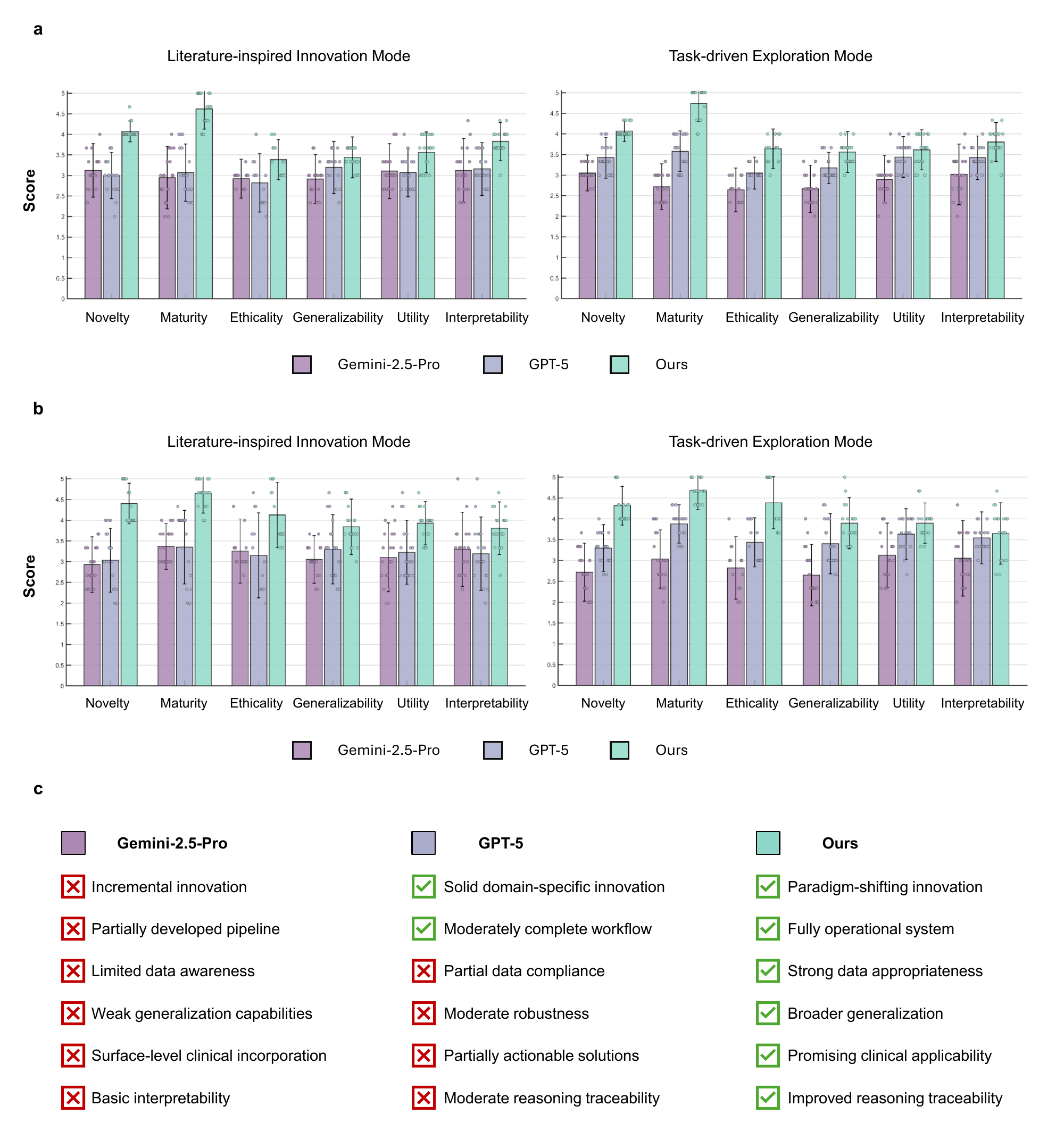}
    \caption{Medical AI Scientist surpasses commercial LLMs in idea generation under combined LLM-based and blinded human evaluation. Models generated research ideas that were anonymized and assessed by three independent experts using a five point scale. a, LLM based evaluation of idea quality. b, Quantitative human assessment across six evaluation criteria. c, Qualitative human analysis of strengths and limitations relative to commercial LLMs.}
    \label{fig:fig3}
\end{figure}

As shown in Fig.~\ref{fig:fig3} a, the Medical AI Scientist consistently outperforms the baselines across six dimensions of idea quality. 
For novelty and maturity, it achieves higher scores in innovation (4.07 vs. 3.00 and 3.12 in literature-based; 4.07 vs. 3.42 and 3.05 in open-ended) and maturity (4.61 and 4.74 vs. \(\leq\)3.58 for the baselines). For technical reliability, it also leads in robustness (3.44 and 3.56 vs. \(\leq\)3.19) and interpretability (3.83 and 3.81 vs. \(\leq\)3.42). Finally, for practical and ethical suitability, the system obtains stronger utility (3.56 and 3.61 vs. \(\leq\)3.44) and ethicality (3.39 and 3.64 vs. \(\leq\)3.05), indicating that the generated ideas are not only more innovative but also more clinically grounded and deployable.

In the human expert assessment (Fig.~\ref{fig:fig3} b), our method consistently achieves the highest scores in technical innovation (4.40 ± 0.49 and 4.32 ± 0.47) and maturity (4.65 ± 0.48 and 4.68 ± 0.47), substantially outperforming GPT-5 and Gemini-2.5-Pro, while also exhibiting lower variance.
This advantage extends to ethicality (up to 4.39 ± 0.63) and robustness (3.90 ± 0.61), where competing models remain below 3.50 on average, indicating more stable and reliable hypothesis generation.
Notably, improvements in utility and interpretability are more moderate (e.g., 3.93 ± 0.53 and 3.81 ± 0.63 in Innovation mode), suggesting that gains in novelty and rigor are accompanied by only incremental advances in practical clarity.
Highlighted by human evaluators’ observations (Fig.~\ref{fig:fig3} c),
our method produces more consistently innovative and mature research ideas, with stronger alignment to clinical relevance and clearer experimental grounding than competing approaches. In contrast, baseline models tend to generate more incremental and less coherent hypotheses, often with higher variability and weaker integration into realistic research workflows.

\begin{figure}
    \centering
    \includegraphics[width=0.94\linewidth]{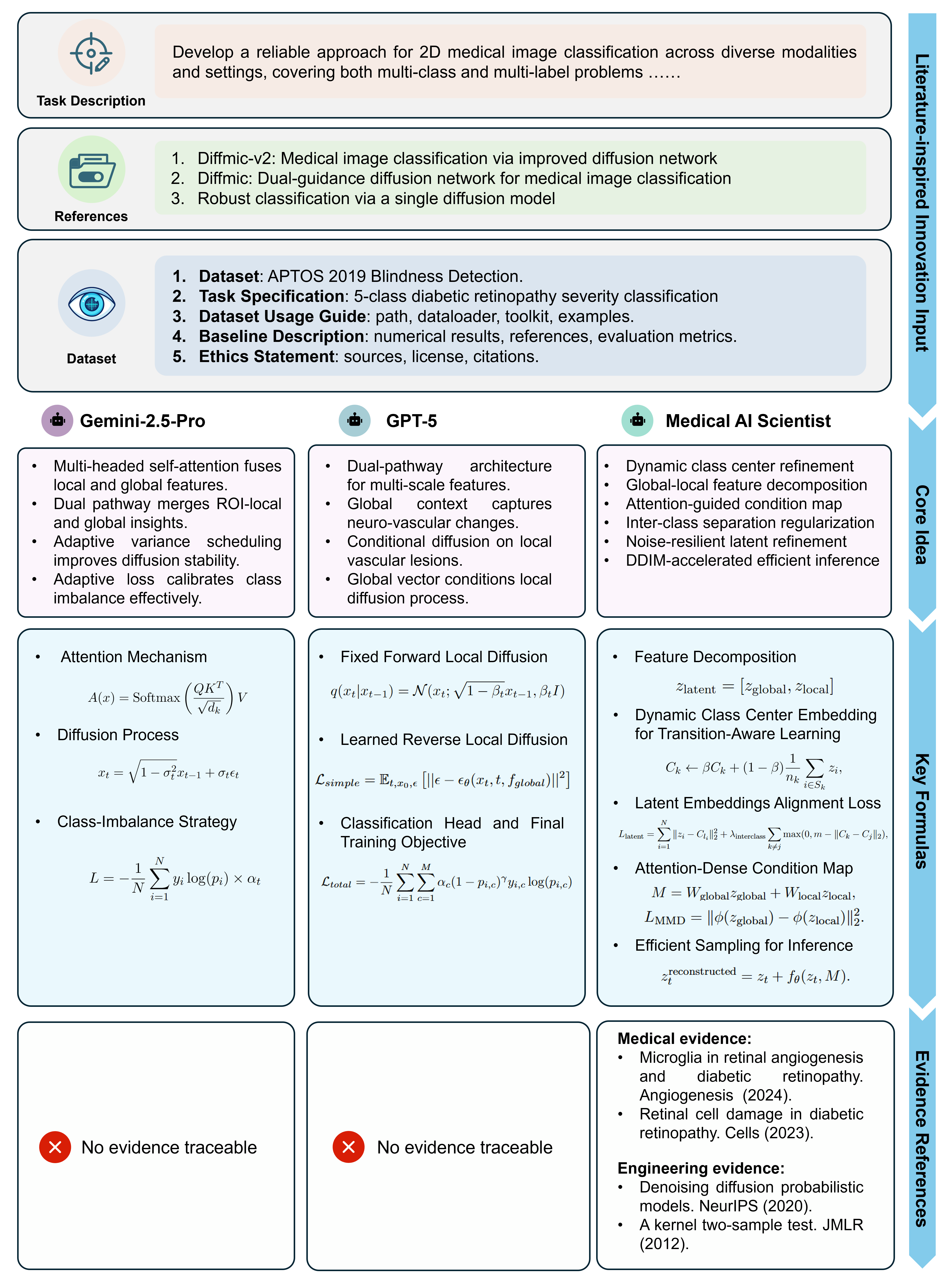}
    \caption{Example of idea generation comparison between Medical AI Scientist and commercial LLMs in the Literature-inspired Innovation Mode.}
    \label{fig:case-study-mode3-1}
\end{figure}

As illustrated in Fig.~\ref{fig:case-study-mode3-1}, this case study compares the idea generation results of our method with those of commercial LLMs under the Innovation mode. All models operate under identical inputs, including the same task description, reference papers, and dataset specification, ensuring a fair comparison.
While commercial models produce reasonable designs, their formulations remain relatively generic and lack strong domain grounding. Their outputs often resemble incremental extensions of prior work, with limited justification from a medical perspective.
In contrast, the proposed method incorporates both medical and engineering evidence into the ideation process, informing model design and learning objectives.
This leads to a more concrete and clinically meaningful formulation, reflected in the richer and more explicit set of equations. 
Consequently, the Medical AI Scientist demonstrates greater implementation detail and improved conceptual novelty, as its designs are guided by disease-related priors rather than abstract extensions of existing approaches.

\subsection{Analysis of experimental implementation}

\subsubsection{Implementation completeness}

\begin{figure}
    \centering
\includegraphics[width=1\linewidth]{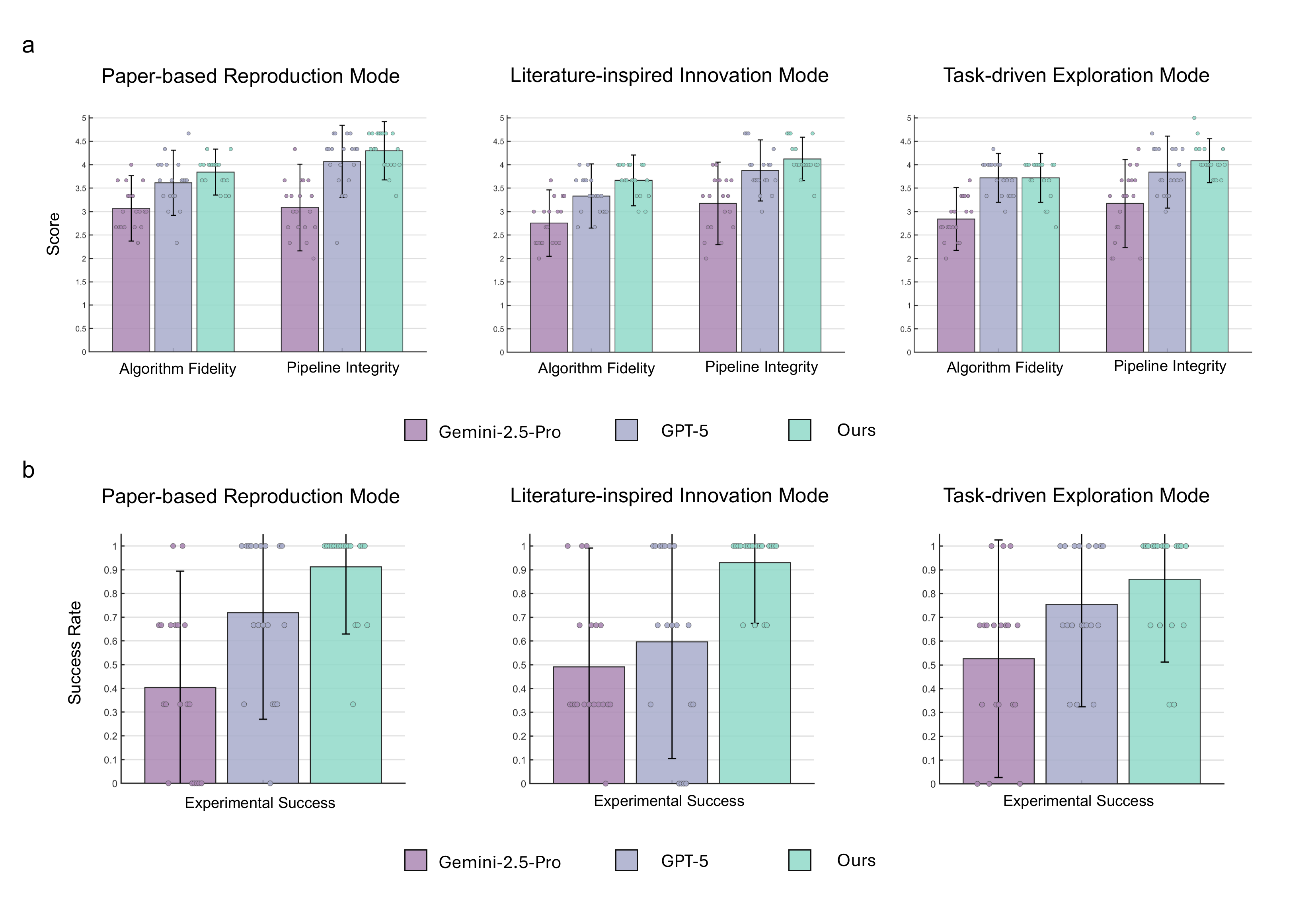}
    \caption{Comparative evaluation of Medical AI Scientist frameworks against commercial LLMs in terms of implementation completeness and experimental success rate. a, Implementation completeness was assessed on a five point scale ranging from 1 to 5. Model generated outputs were anonymized and independently evaluated by two LLM-based judges. b, Experimental success rate measured through quantitative human evaluation.
    }
    \label{fig:figure_2}
\end{figure}

Translating a conceptual research hypothesis into executable code requires preserving methodological coherence between the idea and its technical realization. 
To evaluate this capability, we systematically examined the extent to which finalized research plans were faithfully instantiated in downstream implementations.
As summarized in Fig. \ref{fig:figure_2} a, we quantified experimental success by jointly assessing algorithm fidelity and pipeline integrity, reflecting whether the proposed methodological components were both present and functionally integrated within the resulting codebase.
Across all three experimental modes, our Medical AI Scientist consistently achieved the highest mean scores for both indicators, along with the lowest or near-lowest standard deviations.
In open-ended innovation mode, it reached 3.72 ± 0.52 and 4.09 ± 0.47, respectively, matching GPT-5-Pro while substantially outperforming Gemini-2.5-Pro (2.84 ± 0.67 and 3.18 ± 0.94). The advantage grew clearer in replication mode (3.84 ± 0.49 and 4.30 ± 0.62) and literature-based innovation mode (3.67 ± 0.54 and 4.12 ± 0.46), where our system not only scored highest but also showed the most stable performance.
The results show that the system’s structured refinement process, which couples systematic retrieval from the literature and code repositories with iterative clinician–engineer deliberation, grounds each proposed idea in accessible methodological and technical resources. This integration ensures that finalized research plans are not only scientifically coherent but also practically implementable, with sufficient technical and evidential grounding to enable reliable translation into executable and methodologically faithful code.

\subsubsection{Code execution}

Executing AI-generated research scripts may fail due to unresolved dependencies, dataset incompatibilities, or latent logical errors. These issues become more acute in medical AI research, where heterogeneous clinical data demand specialized preprocessing, domain-specific evaluation metrics, and dedicated software libraries to ensure valid analysis.
To quantify robustness in this context, we measured first-run experimental success across a set of 57 medical AI research instances, comparing experimental results produced by our structured pipeline with those generated directly by the commercial LLM baselines. 

As shown in Fig. \ref{fig:figure_2} b, our approach consistently achieved higher success rates, reflecting the effective resolution of dependency conflicts, enforcement of data compatibility, and runtime-stable logic through iterative refinement and grounding in reference implementations.
By contrast, general-purpose LLM-generated code encountered persistent debugging loops triggered by unresolved runtime errors or became prematurely terminated due to environment configuration issues, preventing successful completion of experiments.
We defined experimental success as stable end-to-end execution of the training pipeline, characterized by successful runtime completion, a decreasing loss trajectory, absence of gradient explosion, and the generation of valid model weight files. Under this definition, our method achieved the highest success rate in all settings, reaching 0.91 in reproduction mode, 0.93 in literature-based innovation mode, and 0.86 in open-ended task mode. In comparison, GPT-5 obtained success rates of 0.72, 0.60, and 0.75, while Gemini-2.5-Pro achieved 0.40, 0.49, and 0.53 under the same conditions. These results show that our system consistently maintains a substantially higher end to end experimental execution success rate across increasing task difficulty.

\subsection{Human and automated evaluation of medical research manuscripts drafting}
To evaluate the translational relevance of autonomous medical research under realistic expert scrutiny, we designed a double-blind user study centered on diabetic retinopathy classification from fundus images while preserving the generality of the framework.
We invited ten independent experts with over five years of first-author experience in AI for healthcare to assess a curated set of 20 manuscripts, including both autonomously generated studies and high-impact human-authored papers. These human-authored works were sampled from leading venues, including the International Conference on Medical Image Computing and Computer Assisted Intervention (MICCAI), International Conference on Bioinformatics and Biomedicine (BIBM), and The IEEE International Symposium on Biomedical Imaging (ISBI).
In parallel, all manuscripts were independently evaluated using the 
\href{https://paperreview.ai/}{Stanford Agentic Reviewer}, an advanced large language model based assessment system, following standardized review criteria aligned with The International Conference on Learning Representations (ICLR) guidelines.

\begin{figure}
    \centering
    \includegraphics[width=1\linewidth]{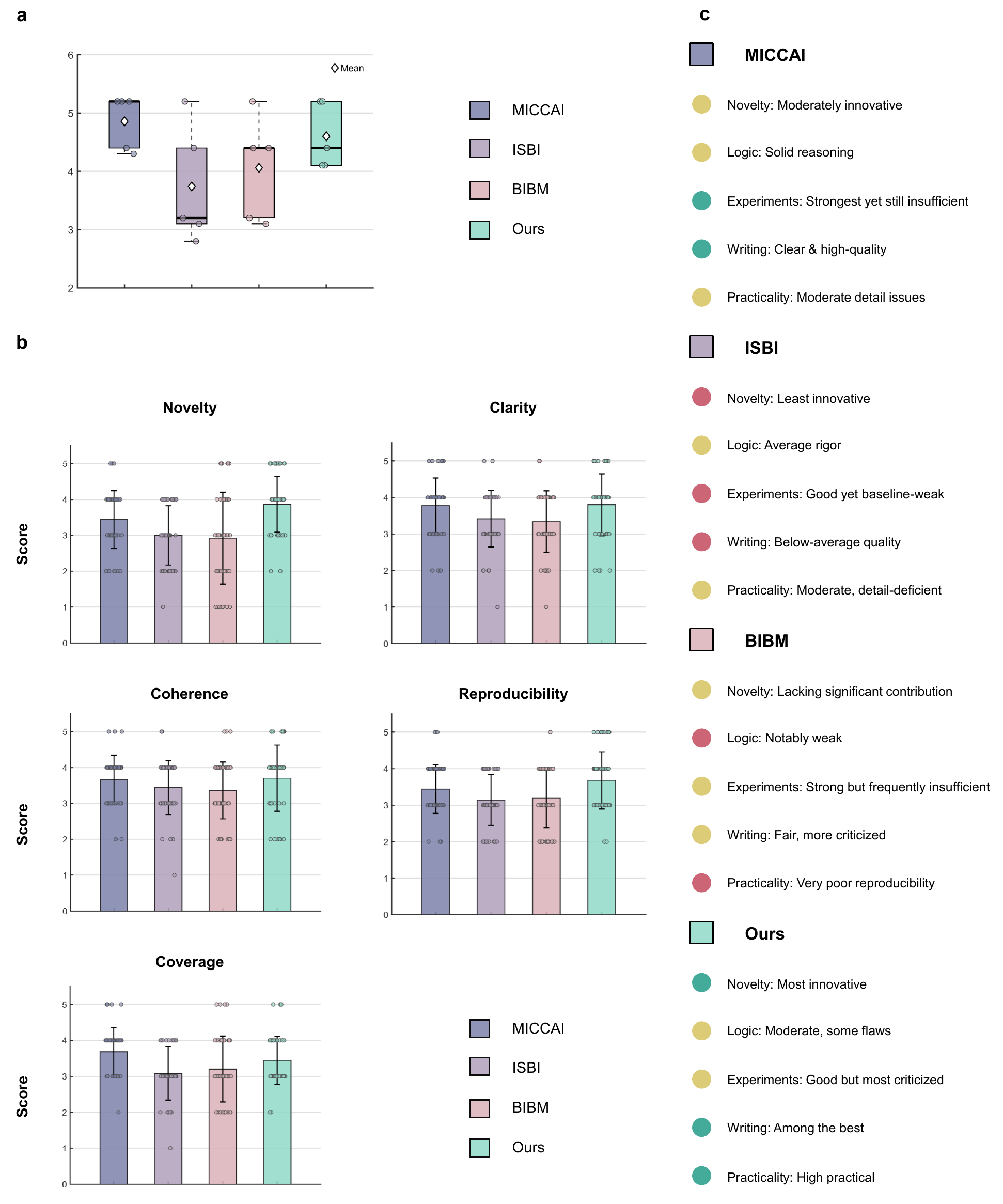}
    \caption{
Anonymized comparison of paper quality on an identical medical task.
Manuscripts generated by Medical AI Scientist achieve performance comparable to MICCAI, ISBI, and BIBM under consistent double-blind evaluation across both quantitative and qualitative assessments:
a, Stanford Agentic Reviewer automatic evaluation. b, Double-blinded scoring (1–5) by 10 medical experts (PhD/postdoc) across five review dimensions. c, Experts’ observations on strengths and limitations.}
    \label{fig:fig4}
\end{figure}

From the AI-based evaluation in Fig. \ref{fig:fig4}a, our method achieves a mean score of 4.60 ± 0.56, comparable to the range observed across representative MICCAI (4.86 ± 0.47), ISBI (3.74 ± 1.02), and BIBM (4.06 ± 0.89) submissions.
According to the double-blind human evaluations in Fig. \ref{fig:fig4}b, our manuscripts demonstrate consistently strong performance across all five dimensions, with scores broadly aligned with those reported for MICCAI, ISBI, and BIBM. In particular, they show competitive results in Novelty, Reproducibility, Coherence, and Clarity, while exhibiting a modest gap in Coverage (3.44 ± 0.67 vs 3.68 ± 0.68), likely reflecting a more focused emphasis on methodological innovation rather than extensive dataset coverage and baseline comparisons.
Qualitative observations (Fig. \ref{fig:fig4}c) from domain experts further highlight the novelty, practical relevance, and clarity of presentation in our manuscripts, alongside solid mid-range assessments in logical coherence and experimental design, with relatively few critical weaknesses noted across comparisons with MICCAI, ISBI, and BIBM submissions.
Overall, these results suggest that our manuscripts achieve a level of quality comparable to that observed across leading venues such as MICCAI, ISBI, and BIBM, as assessed under consistent double-blind evaluation criteria.
We also demonstrated the advantage of our system over other AI-scientist systems by having a manuscript it generated accepted by ICAIS 2025~\citep{icais2025}, which received 114 submissions and had an acceptance rate of 36.8\%.

\subsection{Case study of autonomous medical research process}

\subsubsection{Mode 2: Literature-inspired innovation for medical image classification}

As shown in Fig.~\ref{fig:case-study-mode2}, we further evaluated the proposed automated medical research system to assess its capacity to enrich generated research ideas with medically grounded priors and concrete engineering specifications through its medical–engineering discussion module. Using diabetic retinopathy severity grading as a representative task, the system operated without explicit design instructions and relied solely on reference literature and publicly available codebases. The system demonstrated structured co-reasoning between clinical evidence and implementable methodology: clinical insights from ophthalmic literature motivated the explicit separation of global neurodegenerative context and local vascular pathology, which were subsequently translated into a dual-pathway diffusion-based architecture with imbalance-aware objectives and realizable training protocols. Each design choice was justified by identifiable gaps in prior work and mapped to existing implementations, yielding a hypothesis that was both clinically interpretable and experimentally executable. Quantitative evaluation confirmed that the resulting model achieved competitive performance on imbalanced disease stages, supporting the validity of the underlying reasoning process. Taken together with the paradigm-transfer case study, these results demonstrate that the system can not only identify and adapt novel AI paradigms for specified medical tasks, but also systematically refine them through medical–engineering co-reasoning into fully specified, experimentally validated research hypotheses.

\subsubsection{Mode 3: Task-driven discovery for medical video restoration}

As presented in Fig.~\ref{fig:case-study-mode3}, we evaluated the proposed automated medical research system on a clinically motivated task of restoring high-resolution and temporally consistent endoscopic video from low-quality recordings, thereby assessing its ability to autonomously translate emerging AI paradigms into executable solutions for medical research.
Starting from a minimally specified task description, the system independently grounded the problem in relevant clinical and technical literature, identified temporal inconsistency as a critical unmet requirement, and selected a recently developed continuous-time video restoration paradigm with demonstrable transfer potential. Without manual intervention, this paradigm was adapted to the endoscopic setting through task-specific architectural and training modifications, yielding a complete research hypothesis and an implementable model. The resulting system was experimentally validated through structured ablations and quantitative evaluation, achieving substantial performance gains over a strong baseline. This case study demonstrates that the proposed framework can automatically operationalize novel AI paradigms for concrete medical tasks, progressing from task specification to validated experimental results, and thereby supports its role as a general-purpose engine for automated medical research rather than a task-specific algorithmic contribution.

\section{Discussion}
\label{sec:discussion}
\subsection{Key findings}
In this study, we introduce Medical AI Scientist, an agentic framework that enables end-to-end automation of medical AI research, spanning hypothesis generation, experimental validation, and manuscript composition. By integrating an Idea Proposer, an automated experimental executor, and a hierarchical Manuscript Composer, the system provides a unified solution for the full research lifecycle. A central design feature lies in the clinician–engineer co-reasoning mechanism, which grounds hypothesis generation in verifiable medical evidence and reduces hallucinations. In parallel, the execution module ensures reliable and iterative model development across heterogeneous clinical data, while the manuscript component translates outputs into structured, evidence-based scientific narratives with embedded ethical compliance. To support systematic evaluation, we further introduce Med-AI Bench, a comprehensive benchmark that standardizes assessment across diverse medical research tasks, modalities, and difficulty levels.

Compared with existing approaches to automated scientific discovery, Medical AI Scientist addresses several key limitations. First, general-purpose language models, although capable of generating plausible research ideas, frequently suffer from insufficient grounding in domain-specific evidence, leading to unreliable or non-actionable hypotheses. Second, existing automation frameworks rarely account for the complexity of clinical data formats and the stringent requirements of medical research reporting and ethics. By contrast, our framework unifies these components into a coherent pipeline, ensuring that each stage is both technically rigorous and clinically grounded.

Our experimental results highlight three principal findings. (1) Superior research idea quality: across six evaluation dimensions, the proposed system consistently outperforms commercial language models and approaches human expert-level assessments, demonstrating strong novelty, feasibility, and interpretability. (2) Robust experimental execution: the system achieves high alignment between proposed methods and implemented experiments, with substantially improved success rates in generating executable and self-consistent pipelines for medical AI development. (3) High-quality manuscript generation: under double-blind expert evaluation, generated manuscripts achieve competitive scores relative to top-tier conference publications, with strong performance in coherence, clarity, and reproducibility, and only minor limitations in content coverage. The acceptance of a system-generated manuscript at ICAIS 2025 further provides early evidence of real-world scientific validity.

The broader implications of Medical AI Scientist extend beyond performance gains to a fundamental shift in how medical AI research may be conducted. By significantly reducing the time and expertise required to move from idea to validated results and polished manuscripts, the framework has the potential to accelerate scientific discovery in healthcare. Its ability to systematically explore complex model designs and translate them into executable implementations suggests a complementary role alongside human researchers, particularly in tasks that demand extensive iteration and technical integration. In clinical and translational settings, such a system could lower barriers to innovation, enabling wider participation in medical AI development and fostering more rapid dissemination of clinically relevant solutions.

\subsection{Limitations and future work}

Although our Medical AI Scientist demonstrates promising empirical behavior, several limitations remain before it can be considered to match the best human-produced science.
First, the conceptual design of the method can at times become overly intricate. This complexity not only increases the difficulty of faithful implementation, but also introduces instability during execution. When the intended pipeline proves too demanding, the implementation may implicitly simplify or degrade certain components, leading to deviations from the original design and potentially undermining performance. Second, the depth of experimental evaluation is still limited. Current experiments are conducted strictly on predefined datasets, without sufficient exploration of cross-domain or out-of-distribution scenarios. Finally, despite achieving reasonable performance, the generated method does not yet reach state-of-the-art levels. This gap suggests that further refinement is needed, both in terms of algorithmic design and experimental validation, before the AI-generated approach can be considered competitive with leading methods in the field.

Future work will focus on strengthening the experimental pipeline to enable more comprehensive and rigorous evaluations, thereby improving both the robustness and performance of the method. In parallel, we aim to enhance the quality and expressiveness of visualizations, including both empirical plots and framework illustrations, so as to better communicate the underlying mechanisms and results. Through these efforts, we expect the method to evolve into a more reliable and well-rounded system with stronger empirical competitiveness and clearer presentation.

\section{Methods}
\label{sec:methods}
\subsection{Building an autonomous AI scientist for medical research}

\begin{figure}
    \centering
    \includegraphics[width=1\linewidth]{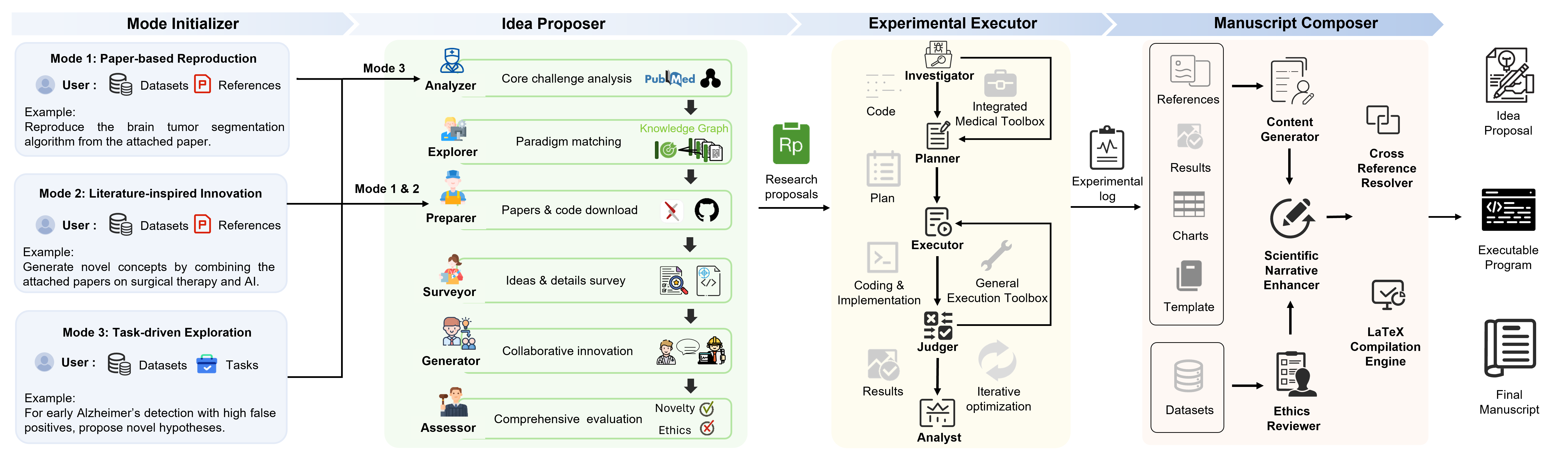}
    \caption{The conceptual illustration of the Medical AI Scientist: A comprehensive system of fully-automated agents for end-to-end scientific discovery in clinical medicine.
    The system offers three user interaction modes: Reproduction (reproducing a specified hypothesis), Innovation (innovating from provided literature) and Exploration (autonomously exploring a given research direction), to streamline medical research process.
    The workflow consists of several phases covering automated idea generation, experiment execution, manuscript writing. 
    }
    \label{fig:framework2}
\end{figure}

As illustrated in Fig.~\ref{fig:framework2}, the Medical AI Scientist comprises three core components: Idea Proposer, Experimental Executor, and Manuscript Composer. 
Each component is implemented as multi-agents that integrate multiple functionalities through carefully designed prompting strategies. The overall system operates via coordinated interactions among these agents. All agents are built upon general-purpose large language models, such as GPT-5, which serve as the base models for handling a broad range of tasks.
For both the Reproduction mode and the Innovation mode, the system takes task instructions, dataset information, and reference papers as inputs, which are then processed sequentially by the Preparer and Surveyor, the Generator, and the Assessor.
In contrast, the Exploration mode operates with only task instructions and dataset information as inputs. Building upon the previous two modes, it first introduces an Analyzer and an Explorer to retrieve the medical baseline paper and the novel technological paradigm paper, thereby establishing a sufficient literature foundation for subsequent idea generation.
The resulting structured ideas are then formulated as research plans and passed to the Experimental Executor for empirical validation, after which the experimental outputs are further processed into structured manuscripts, yielding the final paper.
This entire process enforces a continuous reflect-and-refine cycle, ensuring the final research output (including idea proposal, executable program, and final manuscript) is reproducible and responsible.

\subsection{Idea Proposer}

The Idea Proposer operationalizes medical hypothesis generation as a structured, evidence-grounded reasoning process. The system is organized into a set of interacting functional modules, each addressing a critical component of scientific ideation. This design highlights the central contribution: a unified framework that couples structured knowledge retrieval with clinician–engineer co-reasoning to produce hypotheses that are both novel and verifiable.
At a high level, the Idea Proposer transforms loosely specified medical tasks into executable research hypotheses by iteratively refining problem understanding, identifying appropriate paradigms, and grounding designed ideas in external evidence. This process reduces hallucination and mitigates the tendency of language models to produce superficial or non-actionable ideas.

\noindent\textbf{Analyzer.} 
The Medical Task Analyzer formalizes the input problem by identifying its core clinical and technical challenges. Given a user-provided dataset or research objective, this module performs targeted retrieval over peer-reviewed medical and technical literature to construct a structured task representation based on the academic search engine~\cite{priem2022openalex}. This representation encodes disease context, data characteristics, evaluation constraints, and implicit clinical needs. This step anchors subsequent hypothesis generation in real clinical gaps rather than abstract problem descriptions.

\noindent\textbf{Explorer.} 
Building on the structured task representation, the Paradigm Explorer identifies the most suitable emerging computational paradigms to address the extracted challenges. Instead of relying on static knowledge, it performs dynamic retrieval over recent literature and open-source repositories, jointly considering methodological novelty, empirical performance, and implementation maturity.
A key feature of this module is the explicit alignment between problem structure and algorithmic capability. Candidate paradigms are not selected in isolation but are evaluated based on how their inductive biases and design principles match the identified clinical constraints. For each selected paradigm, the system retrieves corresponding high-quality codebases, ensuring that the resulting hypothesis is directly grounded in executable components.

\noindent\textbf{Preparer and Surveyor.} 
To support informed reasoning, the Preparer and Surveyor jointly construct a structured and executable evidence base that links scientific claims to their operational implementations. The Preparer retrieves relevant literature together with associated code artifacts, normalizing them into a unified representation that captures problem formulations, model designs, and experimental protocols.
Inspired by \cite{AI-researcher}, the Surveyor then performs structured synthesis by decomposing each reference into its core conceptual and methodological primitives. Large language models first extract the fundamental research contribution and methodological skeleton while abstracting away domain-specific terminology to reduce surface bias. These abstract directives are subsequently grounded through a multi-agent process that maps them to canonical mathematical formalisms and aligns them with executable code components from open-source repositories.
This design enables the system to reconstruct prior methods as verifiable workflows rather than static descriptions, thereby transforming existing work into modular and recomposable units for reasoning. 
As a result, this module establishes an evidence-grounded substrate for hypothesis construction by explicitly linking theoretical assumptions with their executable implementations.

\noindent\textbf{Generator.} 
Hypothesis generation is performed by the Generator through a clinician–engineer co-reasoning mechanism that integrates clinical insight with computational design. Rather than relying on unconstrained synthesis, the Generator constructs candidate hypotheses by aligning task-specific challenges with the capabilities of selected paradigms, guided by the structured evidence base. Clinical considerations are introduced in the process to ensure relevance and plausibility, while technical refinements are derived through targeted retrieval and adaptation of existing methods. This bidirectional interaction mitigates the risk of superficial novelty and grounds each hypothesis in verifiable evidence, effectively reducing hallucination. Iterative refinement continues until the hypothesis achieves internal coherence across clinical validity, methodological soundness, and implementation feasibility.
This structured process parallels human-led medical hypothesis formation and enables systematic derivation of high-level ideas from clear gaps in existing literature.

\noindent\textbf{Assessor.} 
The final stage evaluates the generated hypothesis through a combination of scientific and ethical criteria. The Assessor examines conceptual consistency, empirical support, and practical executability. In parallel, an explicit ethics check ensures compliance with biomedical research standards. Hypotheses that fail to meet quality thresholds are returned for refinement, while those violating ethical constraints are rejected. This mechanism enforces rigor and accountability, ensuring that only well-supported and responsible ideas proceed to experimental validation.
The resulting hypothesis is formalized as a detailed research plan, which specifies the algorithmic rationale and anticipated evaluation protocols.

\subsection{Experimental Executor}

The experimental executor is formulated as a structured multi-stage pipeline for traceable and self-correcting model development within a secure Dockerized environment. 
Given a research objective, the \noindent\textbf{Investigator} assembles the required codebase together with domain-specific medical toolboxes to ensure compatibility with heterogeneous clinical data, and provides this unified specification to the \noindent\textbf{Planner}, which decomposes it into a structured, machine-interpretable execution protocol with defined inputs and outputs.
The \noindent\textbf{Executor} instantiates this protocol within a controlled environment by constructing the full training and evaluation pipeline, leveraging general-purpose execution toolchains for scalable and stable implementation. 
Resulting logs, intermediate outputs, and quantitative metrics are assessed by the \noindent\textbf{Judger}, which evaluates consistency between intended design and the observed behavior and produces targeted corrective feedback. 
The {Analyst} consolidates validated results into structured records for downstream use. Through iterative feedback and execution-level correction, the system unifies domain-specific medical processing with general execution infrastructures, enabling reliable, iterative, and self-correcting validation under complex clinical settings.

\subsection{Manuscript Composer}
The Manuscript Composer operates within an end-to-end multi-agent framework that transforms substantiated research materials into a typeset-ready paper. 
The \noindent\textbf{Content Generator} first establishes the global structure of the manuscript by leveraging the organizational patterns of the most relevant reference papers, and subsequently develops section level content grounded in evidence from a structured repository of implementations, experimental logs, and quantitative results.
To preserve narrative coherence, concise summaries of previously generated sections are retained and reused as semantic anchors during subsequent drafting. The Generator further aligns narrative and presentation by automatically generating experimental figures from logged results and synthesizing architectural diagrams from method specifications.
By summarizing current conference and journal policies into structured instructions, the \noindent\textbf{Ethics Reviewer} leverages dataset-specific evidence to rigorously report and cite the origin, license, and ethical approval of each dataset to meet publishing requirements.

In parallel, a \noindent\textbf{Scientific Narrative Enhancer} is introduced to counter the tendency of AI generated text to overemphasize procedural detail, refining the manuscript to improve clarity and the scientific storyline while aligning the writing style with task-specific paradigms.
A \noindent\textbf{Cross-Reference Resolver} subsequently verifies internal references, including equations, figures, sections, and citations. 
Finally, a self-healing mechanism in \noindent\textbf{Latex Compilation Engine} continuously validates the LaTeX source, interpreting compiler feedback to autonomously correct syntactic or structural errors and ensure reliable compilation without manual intervention.
Together, these components enable the automated generation of coherent, compliant, and publication ready medical manuscripts from heterogeneous research artifacts.

\subsection{Construction of Med-AI Bench}
  \label{sec:Med-AI-Bench}
To enable systematic and reproducible evaluation of the Medical AI Scientist, we constructed Med-AI Bench, a benchmark comprising 171 cases derived from 57 high-quality ground-truth medical research papers. Construction began with the six primary data modalities identified in a scoping review of multimodal AI in medicine~\cite{schouten2025navigating}: (1) medical images, (2) videos, (3) electronic health records (EHR, including structured ICU data), (4) text, (5) physiological signals (e.g., ECG and EEG), and (6) multimodal data.

The tasks for each modality were derived from authoritative domain surveys as follows: medical imaging tasks (classification, prognosis, restoration, segmentation, and registration) from a comprehensive review of AI-driven imaging innovations~\cite{pinto2023artificial}; video-analysis tasks (instrument detection, restoration, workflow recognition, intraoperative risk assessment, and postoperative skill assessment) from a scoping review of AI in medical videos~\cite{king2025use}; EHR tasks (risk prediction and clinical decision support) from a comparative analysis of deep learning architectures for EHR~\cite{solares2020deep}; physiological-signal tasks (disease diagnosis and prognosis) from a review of signal-based healthcare applications~\cite{faust2018deep}; clinical text tasks (report summarization, text-based diagnosis/risk assessment, and biomedical question answering) from a UK-focused clinical NLP survey~\cite{wu2022survey}; and multimodal tasks (multimodal diagnosis and cross-modal report generation) from a dedicated multimodal biomedical AI review~\cite{acosta2022multimodal}.
This structured process yielded 19 distinct tasks.

For each task, three representative papers were retrieved from Google Scholar using task-specific keyword combinations, with explicit prioritization of highly cited works. Each paper was independently scored on five dimensions: Code Availability (presence and usability of public implementations), Venue Quality (prestige ranking of the publication venue), Citations (normalized citation count), Year and Complexity (publication recency weighted by methodological intricacy), and Subjective Human Rating (by domain experts). Papers were subsequently ranked and partitioned into three difficulty tiers (hard, medium, easy; one paper per tier per task) from an AI-implementation perspective. For each paper, three cases were constructed using different input modes. The resulting 171 cases form a stratified benchmark that systematically spans technical and clinical complexity, enabling rigorous assessment of hypothesis generation, implementation fidelity, and manuscript quality. It is worth noting that, to speed up the execution and validation of automated experiments, we performed random subsampling on the dataset.

\subsection{Performance assessment of the Medical AI Scientist}
We evaluated the Medical AI Scientist on Med-AI Bench across four core dimensions: (1) Idea Generation, (2) Implementation Completeness, and (3) Code Execution, against the strongest closed-source models (GPT-5 and Gemini-2.5-Pro) under identical input conditions. All evaluation criteria are scored on a five-point scale ranging from 1 to 5, ensuring consistent and interpretable assessment across all cases.

For Idea Generation, the Idea Proposer and baseline models received equivalent prompts (either literature-derived innovation or autonomous exploration of a user-specified direction) and produced full research proposals. Each proposal was scored by a hybrid evaluator that combined LLM-based metrics with blinded assessments from professional clinical AI scientists. Scoring followed standardized rubrics across six dimensions: Novelty (substantive innovation in medical problem modeling), Maturity (completeness and ease of implementation), Ethicality (responsible handling of medical data and constraints), Generalizability (robustness across devices, populations, and institutions), Utility (potential for real clinical adoption), and Interpretability (alignment with medical reasoning and traceability). Explicit evidence grounding ensured high inter-rater reliability. For Implementation Completeness, the full proposals were fed into the Experiment Executor (our system) or the equivalent code-generation modules of the baselines, producing complete executable programs. LLM-based scoring then assessed two aspects: fidelity of core innovative components and completeness of the pipeline (data preprocessing, training, validation, testing, and logging). Code Execution directly deployed the generated code in a predefined Dockerized environment. Success was defined as the fraction of runs that completed without errors, exhibited monotonically decreasing training loss, and produced valid model weights accompanied by quantitative test results. 

In addition, all Medical AI Scientist-generated manuscripts were submitted to the Stanford Agentic Reviewer under the complete ICLR review protocol. The system returned an overall score on a scale from 0 to 10, together with structured strengths and weaknesses, providing an independent multi-criteria validation of scientific rigor.

\subsection{Human expert evaluation}
To assess real-world usability in a controlled yet ecologically valid setting, we restricted the evaluation to a single classic medical AI task: diabetic retinopathy classification on fundus images, while preserving full methodological generality.

We invited 10 independent human experts, each with more than five years of first-author experience in AI-for-healthcare publications. Using a double-blind protocol, experts rated a total of 20 papers: five papers autonomously generated by the Medical AI Scientist on the constrained task and 15 high-impact human-authored papers (five randomly selected via keyword search from each of the MICCAI, BIBM, and ISBI conferences, prioritized by citation rank). To eliminate any potential source bias from formatting or stylistic templates, all human-authored papers had their original templates, fonts, and layouts removed, with only the core content retained.

Experts scored every paper on five dimensions using the same standardized Likert-scale rubrics: Novelty (degree of methodological innovation relative to prior art), Coherence (logical flow and internal consistency of the scientific narrative), Coverage (comprehensiveness of experimental design), Clarity (precision and conciseness of exposition), and Reproducibility (sufficiency of methodological detail). All evaluation criteria are scored on a five-point scale ranging from 1 to 5. Also, all ratings were collected anonymously to eliminate source bias, enabling direct quantitative comparison of perceived quality and practical utility between AI-generated and human-authored medical research outputs.


\section{Related Work}
\label{sec:related}
\subsection{AI agent systems and multi-agent collaboration} The evolution of AI agent systems has shifted from single-agent tool integrations to advanced multi-agent architectures that enable sophisticated collaboration and task decomposition. Early approaches focused on enhancing individual agents' capabilities, such as ReAct~\cite{yao2022react}, which combines reasoning and action by prompting LLMs to generate interleaved thoughts and actions for dynamic environmental interactions. Building on this, Toolformer~\cite{schick2023toolformer} enables LLMs to learn tool usage autonomously through fine-tuning with API calls, supporting zero-shot applications in tasks requiring external resources. These foundations have paved the way for more integrated frameworks like LangChain~\cite{langchain2023}, which facilitates chaining components for complex applications, and its extension LangGraph~\cite{langgraph}, which introduces graph-based orchestration for managing stateful multi-agent systems. Similarly, Semantic Kernel~\cite{semantic-kernel} integrates plugins for enterprise-level AI orchestration with an emphasis on semantic planning and memory persistence.

Building upon these integrated frameworks, advancements in multi-agent collaboration have produced systems that simulate team-based dynamics through role assignments and structured interactions. MetaGPT~\cite{metagpt} employs standardized operating procedures to coordinate agents in workflows akin to software development teams, while CAMEL~\cite{camel} uses role-playing to align autonomous agents with user goals. More specialized frameworks like CrewAI~\cite{crewai} assemble agent teams for sequential tasks such as research synthesis, and OpenAgents~\cite{xie2023OpenAgents} deploys multiple agents to provide accessible capabilities for data analysis, plugin usage, and web navigation. Extending these paradigms, systems like Auto-GPT~\cite{autogpt2023} and Devin~\cite{cognition2024devin} operate as autonomous AI engineers for full-cycle software development, while Manus~\cite{manus2025} and its open-source counterpart OpenManus~\cite{openmanus2025} support complex, cloud-based task execution.
However, these frameworks highlight the evolution toward robust coordination but often lack the deep reasoning required for scientific innovation, such as hypothesis formulation and domain-specific adaptation.

\subsection{Autonomous AI-driven scientific discovery systems} Autonomous scientific discovery systems automate key research stages, from ideation to disseminationd. The AI Scientist~\cite{ai-scientist} pioneers an end-to-end automated pipeline that generates ideas, runs experiments, and drafts manuscripts, operating in an open-ended loop to build upon its own findings. Its successor, AI Scientist-v2~\cite{ai-scientistv2}, enhances this autonomy by incorporating an agentic tree-search for deeper hypothesis exploration and successfully generating a manuscript that passed peer review at a major conference workshop. AI-Researcher~\cite{AI-researcher} introduces a multi-agent architecture that maintains coherence through bidirectional mappings between mathematical concepts and code, mitigating hallucinations. DeepScientist~\cite{deepscientist} frames scientific discovery as a Bayesian Optimization problem, using an agent to iteratively balance exploration and exploitation to discover novel methods. Agent Laboratory~\cite{schmidgall2025agentlaboratory} extends this by automating the execution and reporting of user-provided ideas, acting as an accelerator for human researchers rather than an independent ideator. In contrast, Google's AI co-scientist~\cite{aicoscientist2025} operates as a collaborator in a "scientist-in-the-loop" paradigm, leveraging models like Gemini to assist domain experts with hypothesis generation.

Alongside these frameworks, complementary toolkits have been developed to support AI agent systems by enhancing resource integration and accessibility. ToolUniverse~\cite{gao2025democratizing} provides an expansive repository of scientific tools governed by a standardized AI-tool interaction protocol, enabling agents to discover and orchestrate diverse tools seamlessly. Paper2agent~\cite{miao2025paper2agent} transforms research papers into executable agents by encapsulating their contributions into a standardized Model Context Protocol (MCP), allowing for interactive, natural-language-based reproduction and analysis. Code2MCP~\cite{ouyang2025code2mcp} further streamlines this by converting code repositories into standardized services, facilitating seamless tool incorporation into agent workflows.
While effective for general research automation, these systems frequently overlook clinical necessities such as ethical compliance and specialized data processing, shortcomings our framework mitigates with dedicated medical stages and verification processes.

\subsection{AI applications and challenges in clinical medicine}

Artificial intelligence has made substantial impacts in clinical medicine, with specialized models achieving expert-level performance on well-defined tasks. These tasks include disease classification~\cite{esteva2017dermatologist,kermany2018identifying,rajpurkar2017chexnet}, lesion segmentation~\cite{isensee2018nnunet,hatamizadeh2022unetr}, prognostic prediction~\cite{mobadersany2018predicting,wang2024pathology,chen2024metabolomic} and enhanced surgical navigation~\cite{shvets2018automatic,twinanda2017endonet,maierhein2022surgicaldatascience}.
As technology has evolved, multimodal large language models (MLLMs) have emerged, integrating diverse data types such as text and images to perform more complex, comprehensive tasks. For instance, models like Med-Gemini~\cite{medgemini} leverage vision-language processing to support medical report generation and treatment recommendations, while frameworks such as LLaVA-Med~\cite{li2023llava} facilitate multimodal analysis in radiology.

However, these advances primarily consist of specialized models whose operation and integration still rely heavily on human experts to drive the entire research project. Researchers must be responsible for identifying clinical problems, formulating hypotheses, designing experiments, and ensuring ethical compliance. To our knowledge, no existing framework bridges the autonomous orchestration capabilities of a general AI Scientist with the domain-specific knowledge, tools, and ethical constraints of clinical medicine. Medical AI Scientist aims to fill this gap, enabling autonomous, clinically meaningful, and ethically responsible innovation.




\clearpage
\bibliography{main}

\newpage
\appendix
\label{sec:appendix}

\noindent \textbf{\LARGE{Appendix}}\\

\renewcommand{\thefigure}{A.\arabic{figure}}
\setcounter{figure}{0}

\begin{figure}[!htb]
    \centering
    \includegraphics[width=0.82\linewidth]{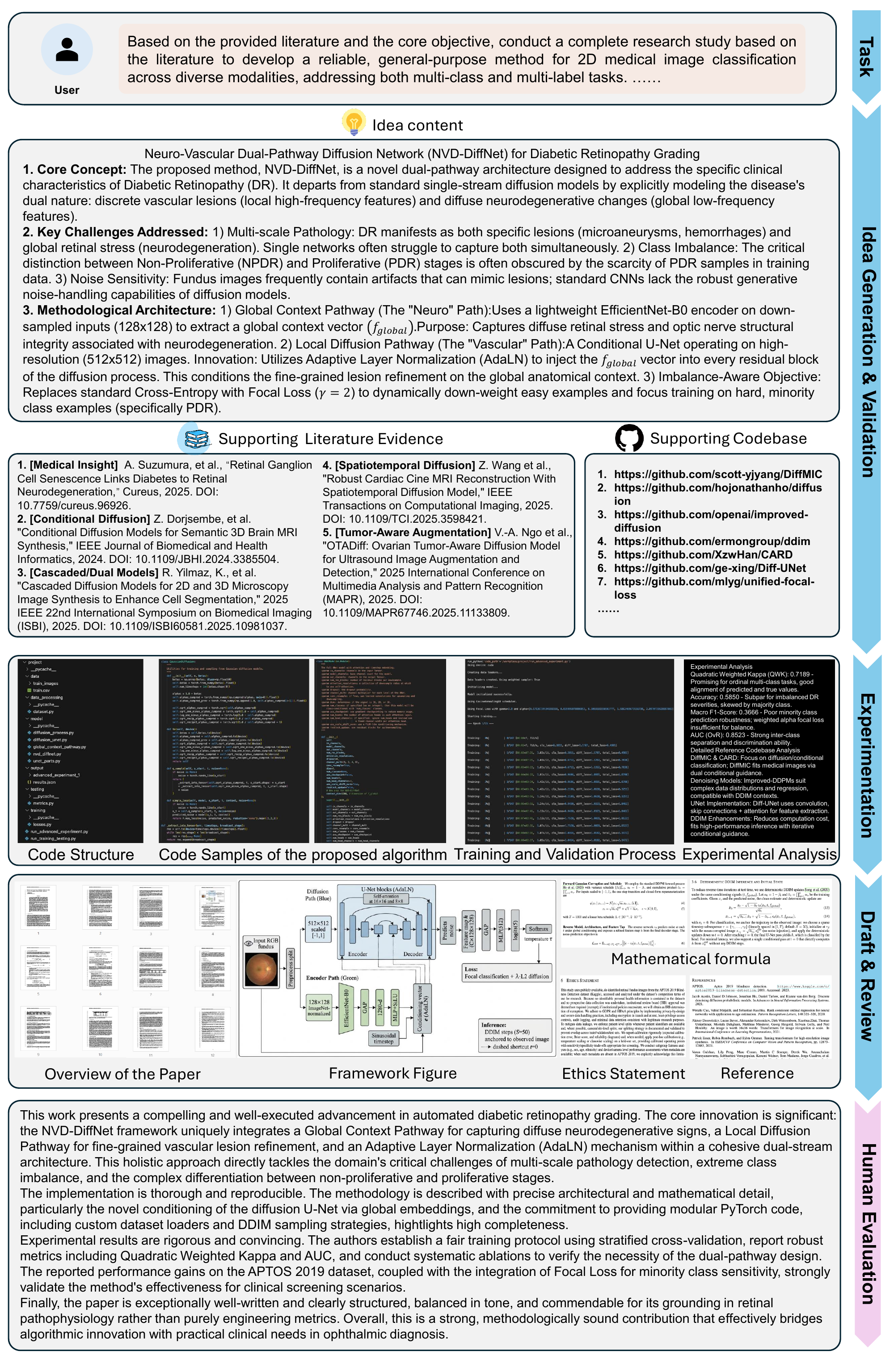}
    \caption{Example of Medical AI Scientist Innovation Mode on the medical image classification task. All necessary processes and outcomes are presented, with human expert's notes in bottom boxes.}
    \label{fig:case-study-mode2}
\end{figure}

\begin{figure}[!htb]
    \centering
    \includegraphics[width=0.94\linewidth]{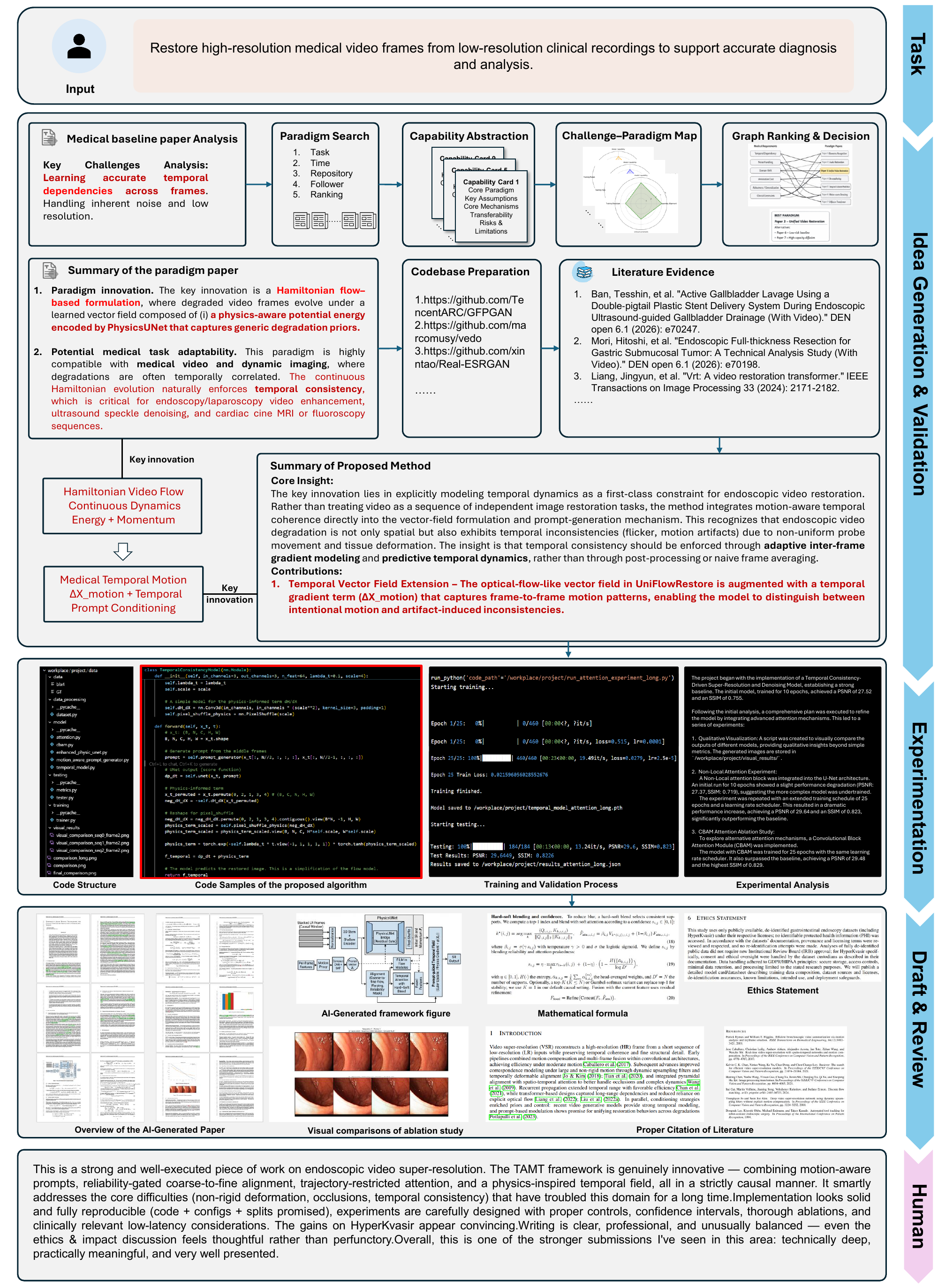}
    \caption{Example of Medical AI Scientist Exploration Mode on the medical video restoration task. All necessary processes and outcomes are presented, with human expert's notes in bottom boxes.}
    \label{fig:case-study-mode3}
\end{figure}

\end{document}